\documentclass[preprint,12pt]{elsarticle}

\usepackage{amssymb}
\usepackage{amsmath}
\usepackage{amsthm}

\usepackage{amsmath,amsfonts}
\usepackage{algorithmic}
\usepackage{array}
\usepackage[caption=false,font=normalsize,labelfont=sf,textfont=sf]{subfig}
\usepackage{textcomp}
\usepackage{stfloats}
\usepackage{url}
\usepackage{verbatim}
\usepackage{graphicx}
\usepackage{colortbl}
\usepackage{balance}

\usepackage{booktabs}
\usepackage{algorithm}
\usepackage[switch]{lineno}
\usepackage{booktabs}
\usepackage{makecell}
\usepackage{multirow}
\usepackage{cancel}
\usepackage{xcolor}
\usepackage{nicematrix}
\usepackage{listings}
\usepackage{hyperref}
\usepackage{xspace}

\usepackage{color}

\definecolor{codegreen}{rgb}{0,0.6,0}
\definecolor{codegray}{rgb}{0.5,0.5,0.5}
\definecolor{codepurple}{rgb}{0.58,0,0.82}
\definecolor{backcolour}{rgb}{0.95,0.95,0.92}

\lstdefinestyle{mystyle}{
    backgroundcolor=\color{backcolour},   
    commentstyle=\color{codegreen},
    keywordstyle=\color{magenta},
    numberstyle=\tiny\color{codegray},
    stringstyle=\color{codepurple},
    basicstyle=\ttfamily\footnotesize,
    breakatwhitespace=false,         
    breaklines=true,                 
    captionpos=b,                    
    keepspaces=true,                 
    numbers=left,                    
    numbersep=5pt,                  
    showspaces=false,                
    showstringspaces=false,
    showtabs=false,                  
    tabsize=2
}

\journal{Neurocomputing}

\begin{document}

\begin{frontmatter}



\title{
Deep Feature Response Discriminative Calibration
}


\author[label1,label3,label4]{Wenxiang Xu}
\author[label1,label3]{Tian Qiu}
\author[label1,label3]{Linyun Zhou}
\author[label1,label3]{Zunlei Feng}
\author[label1,label3]{Mingli Song}
\author[label2]{Huiqiong Wang}
\ead{huiqiong\_wang@zju.edu.cn}

\affiliation[label1]{organization={Zhejiang University},
                addressline={State Key Laboratory of Blockchain and Data Security}, 
                city={Hangzhou},
                postcode={310027}, 
                state={Zhejiang},
                country={China}}
\affiliation[label2]{organization={Zhejiang University},
                addressline={Ningbo Research Institute}, 
                city={Hangzhou},
                postcode={310027}, 
                state={Zhejiang},
                country={China}}
\affiliation[label3]{organization={Zhejiang University},
                addressline={Hangzhou High-Tech Zone (Binjiang) Institute of Blockchain and Data Security}, 
                city={Hangzhou},
                postcode={310027}, 
                state={Zhejiang},
                country={China}}
\affiliation[label4]{organization={Industrial and Commercial Bank of China},
                addressline={Software Development Center}, 
                city={Hangzhou},
                postcode={310012}, 
                state={Zhejiang},
                country={China}}

\begin{abstract}
Deep neural networks (DNNs) have numerous applications across various domains. Several optimization techniques, such as ResNet and SENet, have been proposed to improve model accuracy. These techniques improve the model performance by adjusting or calibrating feature responses according to a uniform standard. However, they lack the discriminative calibration for different features, thereby introducing limitations in the model output. Therefore, we propose a method that discriminatively calibrates feature responses. The preliminary experimental results indicate that the neural feature response follows a Gaussian distribution. Consequently, we compute confidence values by employing the Gaussian probability density function, and then integrate these values with the original response values. The objective of this integration is to improve the feature discriminability of the neural feature response. Based on the calibration values, we propose a plugin-based calibration module incorporated into a modified ResNet architecture, termed Response Calibration Networks (ResCNet). Extensive experiments on datasets like CIFAR-10, CIFAR-100, SVHN, and ImageNet demonstrate the effectiveness of the proposed approach.
\end{abstract}




\end{frontmatter}



\section{Introduction}

Deep Neural Networks are the prevailing architectures for visual representation learning.
The introduction of a wide range of network architectures, including 
AlexNet~\cite{2012ImageNet}, 
VGG-Net~\cite{2014Very}, 
GoogLeNet~\cite{2014Going}, 
ResNet~\cite{2016Deep}, 
DenseNet~\cite{2017Densely}, 
MobileNet~\cite{howard2017mobilenets}, 
and SqueezeNet~\cite{2016SqueezeNet}, 
{as well as theoretical innovations~\cite{guariglia2018harmonic,guariglia2019primality,guariglia2021fractional}, 
has driven neural networks to achieve impressive performance across various fields, such as
image classification~\cite{yang2019hyperspectral,basha2020impact},
object recognition\cite{qureshi2022neurocomputing,zhou2024patchdetector},
signal decomposition~\cite{mallat1989theory,zheng2019framework},
fractal-wavelet modeling~\cite{guido2017effectively,guariglia2016fractional},
mathematical equation solving~\cite{MENG2023PINN,raissi2024forward},}
disaster monitoring~\cite{muhammad2018early,gupta2021deep},
and medical image analysis~\cite{yu2021convolutional,niyas2022medical}.

The quality of the features extracted by the model directly influences the model's performance.
Therefore, optimizing the model through adjustments to neuron response values is a plausible approach for further improvement.
Previous works have also optimized models through feature calibration methods. 
Such techniques include normalization approaches, like Batch Normalization (BN)~\cite{ioffe2015bn}, intermediate network modules, such as Squeeze-and-Excitation Networks (SENet)~\cite{hu2018senet}, {and optimizers, such as Lion~\cite{chen2024symbolic} and Gradient Search-based Binary Runge Kutta Optimizer (GBRUN)~\cite{dou2024gbrun}.} These methods have enhanced model performance by refining features through channel and spatial attention mechanisms.

While these methods have achieved notable improvements, they primarily focus on scaling raw feature values without sufficiently addressing feature distinctiveness.
{In contrast, our proposed response value calibration technique introduces additional calibration values to the extracted features with the aim of enhancing their distinctiveness. This method diverges from existing techniques by emphasizing the optimization of feature distribution and enhancing their distinctiveness rather than just scaling feature values. By focusing on feature distribution and distinctiveness, our approach offers a novel contribution to the field.}

Specifically, we propose a response value calibration technique that introduces additional calibration values to features extracted by deep learning models, thereby enhancing their discrimination.
These calibration values essentially compute the confidence of the original features based on the integral of the Gaussian probability density function.
More precisely, we fit each neuron with mean and variance parameters to model the Gaussian distribution that the neuron's responses follow. 
Using the probability density function of this distribution, we calculate a confidence value for each response: the confidence for a response value below the mean corresponds to the integral of the probability density function up to that value, while the confidence for a response above the mean is given by the integral from the symmetrical point of that value with respect to the mean.
Based on this response value calibration scheme, we introduce the Response Calibration Layer (RC Layer), similar to the SE layer. Models integrated with this calibration layer are termed Response Calibration Networks (ResCNet).

In summary, the main contributions of this paper are as follows:
\begin{itemize}
    \item Introducing a novel approach to calibrate neuron response values from a perspective of discriminative distribution correction. {This method takes a distribution-based approach and employs probability density functions to compute additional values that increase the specificity of the original features.}
    \item A plug-in-based calibration module is devised. Based on this plugin, we developed a new model architecture called ResCNet. {As a modified version of the ResNet model, ResCNet introduces an additional response calibration branch between each convolutional block. This branch offers supplementary calibration values for the model’s features.}
    \item Extensive experiments on CIFAR-10, CIFAR-100, SVHN, and ImageNet datasets demonstrate the effectiveness of the proposed approach.
     The calibrated model not only enhances classification performance but also extracts more discriminative features.
\end{itemize}

\begin{figure*}
    \centering
    \includegraphics{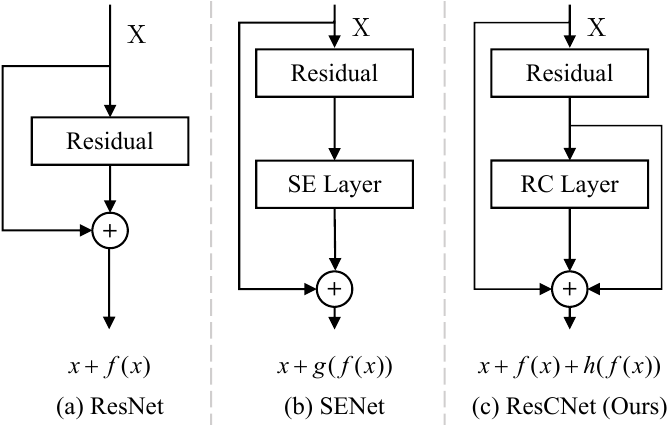}
    \caption{
        \textbf{Module architecture.}
        From the module architecture, it can be seen that SENet further scales the features extracted by the ResNet residual branches, whereas our proposed architecture provides additional calibration of these features.
    }
    \label{fig:first_pic}
\end{figure*}

\section{Related Work}
{The quality of image features is crucial for various visual tasks. Recently, significant progress has been made in adjusting the response values of neurons in deep learning models to enhance the performance of various visual tasks.
These methods include regularization techniques~\cite{ioffe2015bn,ba2016ln} analogous to standardization operations in traditional feature engineering, activation functions~\cite{howard2017mobilenets,hendrycks2016gaussian} that introduce nonlinear transformations to enable neural networks to learn and represent complex nonlinear relationships, and various intermediate modules~\cite{hu2018senet,li2019sknet} based on attention mechanisms.
Furthermore, nature-inspired metaheuristic algorithms~\cite{erdacs2023optimum,sait2024optimal} are gaining popularity due to their ease of application and their ability to avoid local optimum points.}
In this section, we briefly present the feature response adjustment techniques related to this work, which are primarily categorized into regularization methods, activation functions, and attention mechanisms.

\noindent
\textbf{Regularization Methods.}
Training deep neural networks is challenging since the distribution of each layer’s inputs will change with the gradually optimized parameters of the previous layers. This phenomenon is termed internal covariate shift~\cite{ioffe2015bn}.
These issues can result in a slower convergence rate for convolutional neural networks when aiming for the optimal solution.
To alleviate this dilemma, various regularization methods have been proposed, such as \emph{Batch Normalization}~\cite{ioffe2015bn}, \emph{Layer Normalization} (LN)~\cite{ba2016ln}, \emph{Group Normalization} (GN)~\cite{wu2018gn}, and \emph{Instance Normalization} (IN)~\cite{ulyanov2016instance}.
Normalization techniques, by adjusting the responses of neurons, accelerate the model's convergence speed.
Batch Normalization depends on the batch size, whereas Layer Normalization, Group Normalization, and Instance Normalization do not depend on the batch size.

{Although normalization layers have played a positive role in accelerating model training, their potential in model optimization has yet to be fully explored. Currently, the primary function of normalization layers is to standardize the outputs of the preceding layer, but this process does not deeply consider the potential significance of feature values. This work focuses on exploring how to measure and utilize the importance of feature values more thoroughly from the perspective of response distribution, thereby advancing model optimization. }

\noindent
\textbf{Activation Functions.}
Activation functions introduce non-linearity to models by adjusting feature responses, including ReLU, ReLU6~\cite{howard2017mobilenets}, GELU~\cite{hendrycks2016gaussian}, Tanh, and Sigmoid. 
Specifically, ReLU filters out values that are less than 0. 
ReLU6, introduced in MobileNets~\cite{howard2017mobilenets}, is essentially a modification of the rectified linear unit that limits the activation to a maximum size of 6, enhancing the robustness of low-precision computations. 
GELU adjusts neuron responses from the perspective of mathematical expectation.
Additionally, Tanh and Sigmoid map the activation values to specific intervals.

{Although activation functions provide some degree of selective filtering of the original feature values, most activation functions, such as Sigmoid and Tanh, merely map feature values to fixed intervals, while others like ReLU perform simple feature selection. These approaches do not take into account the distribution characteristics of the feature values, thereby overlooking their potential positive impact. In contrast, this work aims to assess the importance of the original feature values more precisely from the perspective of their distribution, seeking to achieve a more precise quantification and utilization of features.}

\noindent
\textbf{Attention Mechanism.} 
There have been several articles on attention mechanisms, such as SENet~\cite{hu2018senet}, SKNet~\cite{li2019sknet}, CBAM~\cite{woo2018cbam}, {Involution~\cite{li2021involution}, and Halo~\cite{vaswani2021scaling}.} SENet {and Involution} intrinsically introduce dynamics conditioned on the input, which can be regarded as a self-attention function on channels. SKNet employs convolution kernels of different sizes and a method similar to SENet to dynamically calculate the weight of each convolution kernel. CBAM {and Halo} combine channel attention and spatial attention, concatenating them to determine the weight of each feature map.

These articles primarily focus on learning a weight at each level of the feature map through the use of attention mechanisms.
{Our approach differs from these techniques in that it directly scales features in the residual branch. This work evaluates the importance of the original feature values more meticulously from the perspective of their distribution and introduces additional adjustment values for the original features.}

\section{Pre-analysis for Activation Responses}\label{sec3}

\begin{figure*}[!t]
    \centering
    \includegraphics[width=\textwidth]{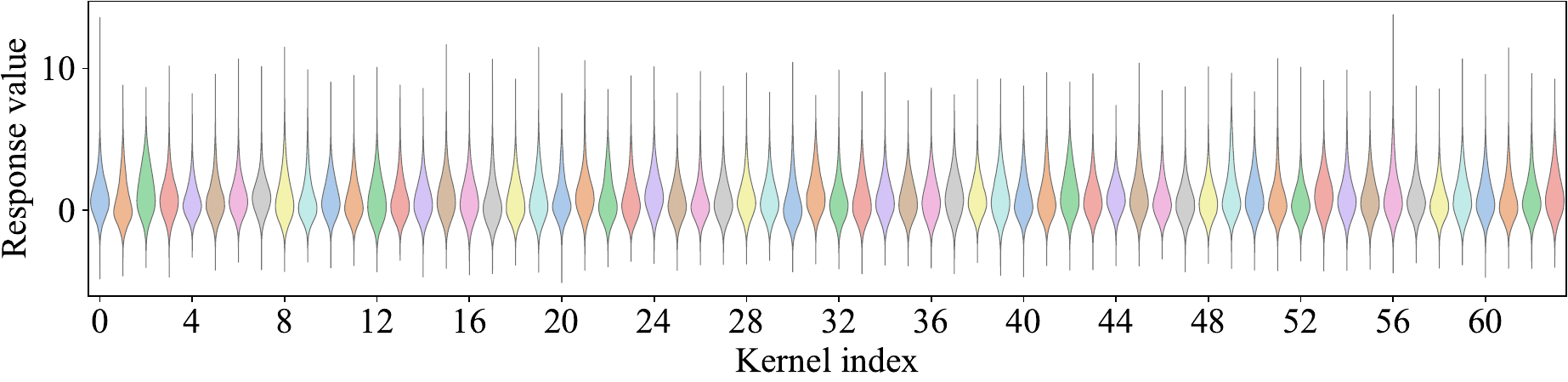}
    \caption{
        \textbf{The distribution of response values 
        after model convergence using ResNet-32 on CIFAR-100 dataset.}
        From the outer contour of the violin plot, it can be observed that the response values of a single convolutional kernel (neuron) roughly follow a Gaussian distribution.
    }
    \label{fig:res32_violin}
\end{figure*}

We first obtained the features of the samples after passing through the feature extraction layer and then performed a global average pooling (GAP) operation to obtain the final feature values. The dimension of these feature values corresponds to the number of output channels from the last convolutional layer. 
For the entire dataset, we obtained a final data dimension of [N, C] (N is the sample size of the entire dataset, and C refers to the number of convolutional kernels in the final convolutional layer). 
Specifically, we use the CIFAR-100 dataset~\cite{krizhevsky2009learning}, and the model employed is ResNet-32.
We visualized the feature values, as shown in Fig.~\ref{fig:res32_violin}.

From the visualized results, as shown in Fig.~\ref{fig:res32_violin}, it can be observed that the response values of each convolutional kernel approximately follow a Gaussian distribution, with different kernels adhering to different Gaussian distributions. 
{It is reasonable to assume that features follow a Gaussian distribution, given the widespread occurrence of Gaussian distributions in both statistics and the natural world.
Specifically, in deep learning models, a neuron's output is typically a weighted sum of multiple input features. The Central Limit Theorem states that when the number of independent random variables is sufficiently large (greater than $30$), the distribution of their sum will approximate a Gaussian distribution, even if the random variables themselves are not Gaussian distributed. According to the Central Limit Theorem, this output distribution is likely to approximate a Gaussian distribution. Furthermore, in modern network architectures, intermediate modules like BN, LN, and activation functions such as GELU are predicated on the assumption that feature values follow a Gaussian distribution. The encoder in Variational Auto-Encoders (VAE)~\cite{kingma2013auto} also relies on this assumption.}
By integrating experimental observations with prior research, and under the assumption that features follow a Gaussian distribution, we developed techniques for adjusting response values and implemented them in a specialized network module.

\section{Self-adaptive Response Calibrator}
In this section, we introduce the core technique of this paper: adjusting the original feature values based on the assumption that they follow a Gaussian distribution. Specifically, we design two trainable parameters, mean and standard deviation, for each convolutional kernel to fit its corresponding Gaussian distribution. Then we use the integral of the Gaussian probability density function corresponding to this distribution to measure the neuron response values, thus obtaining the final calibration values.
The calibration feature values are incorporated into the original feature values during the model's training process.

Based on this response value calibration method, we propose a plugin-based module to integrate the calibration values into mainstream neural networks.
The plugin-based module is incorporated into a modified ResNet architecture, which we call Response Calibration Networks. 

\subsection{Calibration Value Calculation}

In the proposed method, we consider the Gaussian distribution's mean to reflect the convolutional kernel's general capability in processing features, which is of significant importance. 
Response values closer to the mean are assigned larger weights, whereas response values farther from the mean are assigned smaller weights.

It is hypothesized that the response values generated by any convolutional kernel follow a Gaussian distribution $\mathbf{E_{k}} \sim \mathcal{N}(\mathbf{u_{k}}, \boldsymbol{\sigma_{k}^{2}})$, where $ \mathbf{u_{k}} $ and $ \boldsymbol{\sigma_{k}} $ are the mean and standard deviation, respectively, of the convolution kernel $k$.
For an response value $ a_{k} \leq \mu_{k} $, the corresponding weight $ w_{k} $ is calculated as follows:
\begin{equation}
w_{k} = \frac{1}{\sigma_{k} \sqrt{2 \pi}} \int_{-\infty}^{x} \exp \left(-\frac{(t-\mu_{k})^{2}}{2 \sigma_{k}^{2}}\right) dt.
\end{equation}

The above results can be simplified with the Error function (erf):
\begin{equation}
w_{k} = \frac{1}{2}\left[1+\operatorname{erf}\left(\frac{a_{k}-\mu_{k}}{\sigma_{k} \sqrt{2}}\right)\right].
\end{equation}

The formula can be further simplified as follows:
\begin{align}
w_{k} &= \Phi(x) \nonumber \\
    &=\frac{1}{2}\left[1+\operatorname{erf}\left(\frac{{x}}{\sqrt{2}}\right)\right],
\end{align}
where $ x = (a_{k} - \mu_{k}) / \sigma_{k} $ and $\Phi(x) = P(X\le x), X\sim \mathcal{N}(0,1)$ is the cumulative distribution function (CDF) of the standard normal distribution.
This operation ensures that the closer the response value is to the mean, the greater the corresponding weight.

If $ a_{k} > \mu_{k} $, signifying that the response value is distant from the mean, the weight value $w_{k} = 1 - \Phi(x) $.
In other words, we use the cumulative distribution function at the point symmetric to the mean position as the final weight.
This operation reduces the weight associated with a response value as it deviates further from the mean.

To summarize, the weight $ w_{k} $ is defined by the following expression:
\begin{equation}\label{eq:weight_final_eq_v2}
w_{k} = \begin{cases}
1- \Phi(x), &a_{k} > \mu_{k} \\
\Phi(x), &a_{k} \le \mu_{k}
\end{cases}.
\end{equation}

The adjustment value $c_{k}$ is computed as follows:
\begin{align}\label{eq:adjustment_value}
c_{k} &= a_{k} * w_{k} \nonumber \\
      &= \begin{cases}
        a_{k} * (1- \Phi(x)), &a_{k} > \mu_{k} \\
        a_{k} * \Phi(x), &a_{k} \le \mu_{k}
    \end{cases}.
\end{align}

From Eq.~\ref{eq:weight_final_eq_v2}, it can be observed that when the neuron response value is less than the mean, our weight formula aligns with that of GELU~\cite{hendrycks2016gaussian}. Conversely, when the neuron's response value exceeds the mean, our weight values differ from GELU's; however, their total remains 1.

\begin{figure*}[!t]
    \centering
    \includegraphics[width=0.8\textwidth]{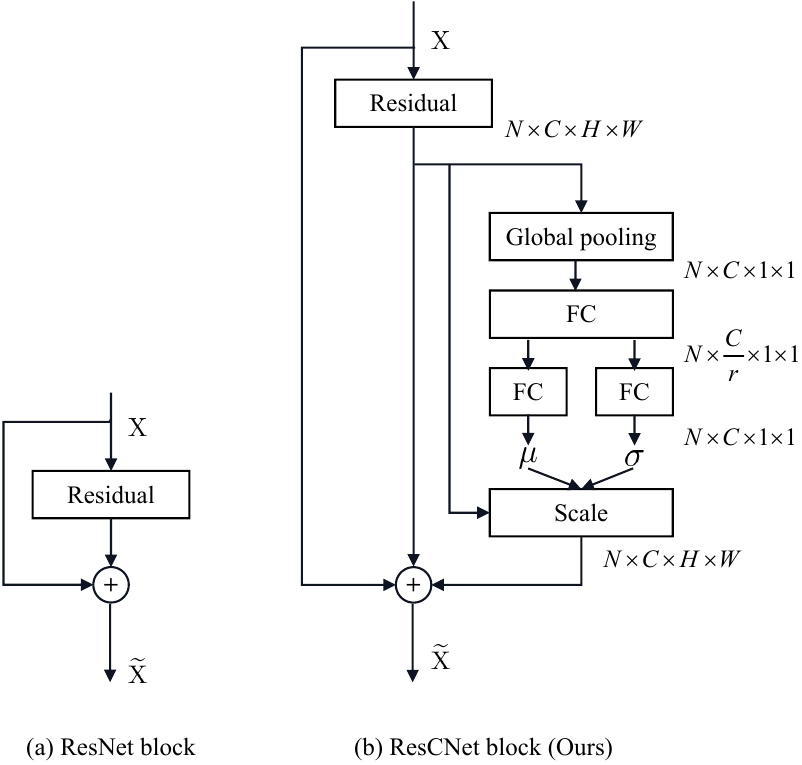}
    \caption{
        \textbf{ResNet block and ResCNet block.}
    }
    \label{fig:rcnet}
\end{figure*}

\subsection{Integration Scheme for Calibration Value}

We apply the adjustment values sequentially before the original activation function. 
Assuming the activation function is ReLU, the adjustment values are incorporated into its input. 
Taking the basic block of a ResNet as an example, the input values enter the basic block and split into two branches: the residual branch and the shortcut branch. The residual branch processes the input in the following order: Conv $\rightarrow$ BN $\rightarrow$ ReLU $\rightarrow$ Conv $\rightarrow$ BN. The output of the residual branch is then added to the shortcut branch and passed through a ReLU function. We integrate 
the proposed method 
(similar to the SE Layer) 
after the second BN layer.
By examining the feature processing flow, we can observe the difference between our method and SE layer. SE layer scales the original features and uses them as the final output of the residual branch. In contrast, 
the proposed method adds the calibration values to the original feature values.

We have developed this scheme as a module, facilitating its integration into mainstream models. 
For instance, we have integrated the calibration module into ResNet, naming this model variant Response Calibration Networks (ResCNets). 
A basic block design is illustrated in Fig.~\ref{fig:rcnet}.
We employ Multi-Layer Perceptrons (MLPs) to fit the mean and standard deviation corresponding to each convolutional kernel. Subsequently, we use Eq.~\ref{eq:adjustment_value} to scale the original feature values, resulting in the calibration values.

\section{Experiments}\label{experiments}

In this section, we conducted extensive experiments to validate the effectiveness of the method proposed in this paper.
We compared the performance of the original ResNet models with other variants, such as SENet, the ResCNet proposed in this paper across various datasets. 
Additionally, in the ablation study section, we also explored other integration methods for calibration values.

\subsection{Experiment Setting}

\subsubsection{Datasets}
The datasets we adopted contain
CIFAR-10,
CIFAR-100~\cite{krizhevsky2009learning},
SVHN~\cite{netzer2011reading},
and ImageNet~\cite{deng2009imagenet},
which are widely used datasets for the classification task.

\noindent{\textbf{CIFAR-10 \& CIFAR-100.}}
Both of the CIFAR datasets~\cite{krizhevsky2009learning} consists of 60,000 colour images of size $32 \times 32$ pixels. 
CIFAR-10 has 10 distinct classes, 
and CIFAR-100 contains 100 classes.
Each dataset is split into a training set with 50,000 images and a test set with 10,000 images.

\noindent{\textbf{SVHN.}}
The Street View House Numbers (SVHN) dataset~\cite{netzer2011reading} contains a total of 630,420 color images with a resolution of $32 \times 32$ pixels. Each image is centered about a number from one to ten.
The official dataset split contains 73,257 training images and 26,032 test images, but there are also 531,131 additional training images available.

\noindent{\textbf{ImageNet.}}
ImageNet is a vast image database widely used in visual object recognition research. 
The most highly-used subset of ImageNet is the ImageNet Large Scale Visual Recognition Challenge (ILSVRC) 2012 image classification dataset. This dataset spans $1,000$ object classes and contains $1,281,167$ training images and $50,000$ validation images.

\begin{table*}[!t]
    \centering
    \resizebox{\textwidth}{!}
    {
\begin{tabular}{@{}l|cc|cc|ccc@{}}
\toprule
 & \multicolumn{2}{c|}{Params (K)} & \multicolumn{2}{c|}{FLOPs (M)} & \multicolumn{3}{c}{Accuracy(\%)} \\
\midrule
Dataset &CIFAR-10 \& SVHN  & CIFAR-100  &CIFAR-10 \& SVHN  & CIFAR-100 &  CIFAR-10  & CIFAR-100 & SVHN\\ \midrule
ResNet-32 &466.91 & {472.76} & {70.37}  & 70.38 & 93.72 & 71.91 &95.82  \\
SENet-32 &480.35 & {486.20} & {70.67}  & 70.68 & 93.73 & 72.26 &95.77 \\
ResCNet-32 (Ours) & {488.33} & {494.18} &{70.68}    & 70.68 & \textbf{93.79}  & \textbf{72.76} & \textbf{95.87}  \\ \midrule
ResNet-56 &855.77 & {861.62} & {127.91}  & 127.92 & 94.51 & 73.79  &96.00 \\
SENet-56 &879.96 & {885.81} & {128.45}  & 128.46 & 94.51 & \textbf{74.68} &95.88  \\ 
ResCNet-56 (Ours) &894.33 & {900.18} & {128.47}   & 128.47 & \textbf{94.61} & 73.81 &\textbf{96.10} \\ \midrule
ResNet-50 &23520.84 & {23705.25}  & {1304.94} & {1308.84} & 95.81 & 80.79 &96.29 \\
SENet-50 &26078.39 & {26262.80} & {1314.48} & {1321.98} & 95.75 & 80.68 &96.26 \\
ResCNet-50 (Ours) &27324.41 & {27508.82} & {1316.25}  & {1316.43} & \textbf{95.83} & \textbf{80.87} &\textbf{96.31}  \\ \bottomrule
\end{tabular}
}
    \caption{
    \textbf{Results of the sequential scheme (ResCNet).}
    Model architectures with 32, and 56 layers have 3 stage blocks (downsampling in the last two stages), and 50 layers has 4 stage blocks (downsampling in the last three stages).
    We compare the performance of ResNet, SENet, and ResCNet models with the same number of layers. 
    The \textbf{bold} indicates the best performance.}
    \label{tab:tab_rcnet}
\end{table*}

\subsubsection{Parameters}
Our study trained convolutional neural networks for 200 epochs on datasets including CIFAR-10, CIFAR-100, and SVHN using the Stochastic Gradient Descent (SGD) optimizer. For the ImageNet dataset, training was conducted for 90 epochs. During the initial phase, models were warmed up for 5 epochs. The learning rates were set as follows: 0.1 for CIFAR-10, CIFAR-100, and ImageNet; 0.01 for SVHN. The batch sizes were 128 for CIFAR-10, CIFAR-100, and SVHN, and 256 for ImageNet.
In terms of learning rate schedulers, aside from using the StepLR scheduler for the ImageNet dataset, all other datasets employ a CosineAnnealingLR~\cite{loshchilov2016sgdr} scheduler.

Regarding the reduction hyperparameter for the sequential scheme, a value of 4 was used for ResNet architectures with 32 and 56 layers, while a value of 16 was applied to ResNet with 18, 34, and 50 layers.

\subsection{Experimental Results}
\subsubsection{Comparision with Other Calibration Methods}
We compared the performance of the original ResNet models with other variants, including SENet and the ResCNet proposed in this paper.
The results on the CIFAR-10, CIFAR-100, and SVHN datasets are shown in Tab.\ref{tab:tab_rcnet}. 
It can be observed that as the depth of the model increases, the proposed ResCNet outperforms both ResNet and SENet in terms of performance.
In scenarios with a larger number of model parameters and deeper layers, the proposed ResCNet performs better than both SENet and ResNet.

\begin{table*}[!t]
    \centering
    \resizebox{\textwidth}{!}{
    \begin{tabular}{l|cccc}
    \toprule
                & Params (M) & FLOPs (G) &  Top-1 Accuracy (\%) & Top-5 Accuracy (\%)\\ \midrule
       ResNet-18 (TorchVision)$^*$    &11.70 &1.81 & 69.758 & 89.078 \\
       ResNet-18 (Re-Implementation)  &11.70 &1.81 & 70.098 & 89.260 \\
       ResCNet-18 (Ours)  &11.82 &1.83 & \textbf{70.354}  & \textbf{89.586}  \\ \midrule
       ResNet-34 (TorchVision)$^*$    &21.80  &3.66 & 73.314 & 91.42 \\
       ResNet-34 (Re-Implementation)  &21.80  &3.66 &73.486  &91.526  \\
       ResCNet-34 (Ours)              &22.04 &3.68 & \textbf{73.816}  & \textbf{91.546}  \\ \midrule
       ResNet-50 (TorchVision)$^*$    &25.60 &4.09 & 76.130 & 92.862 \\
       ResNet-50 (Re-Implementation)  &25.60 &4.09 &76.200 &92.886  \\
       SENet-50$^*$ & - & - &76.710  &\textbf{93.380}  \\
       ResCNet-50 (Ours) &29.36 &4.15 &\textbf{76.920}  &{93.298}  \\ 
       \bottomrule
       
    \end{tabular}}
    \caption{
        \textbf{ImageNet result using ResNet and ResCNet.}
        ``$*$'' indicates that the result is from the original paper or the official result.
        ResNet and ResCNet were trained for 90 epochs, whereas SENet was trained for 100 epochs.
        The \textbf{bold} indicates the best performance.
    }
    \label{tab:imagenet_result}
\end{table*}

\begin{figure*}[!t]
    \centering
    \includegraphics[width=\textwidth]{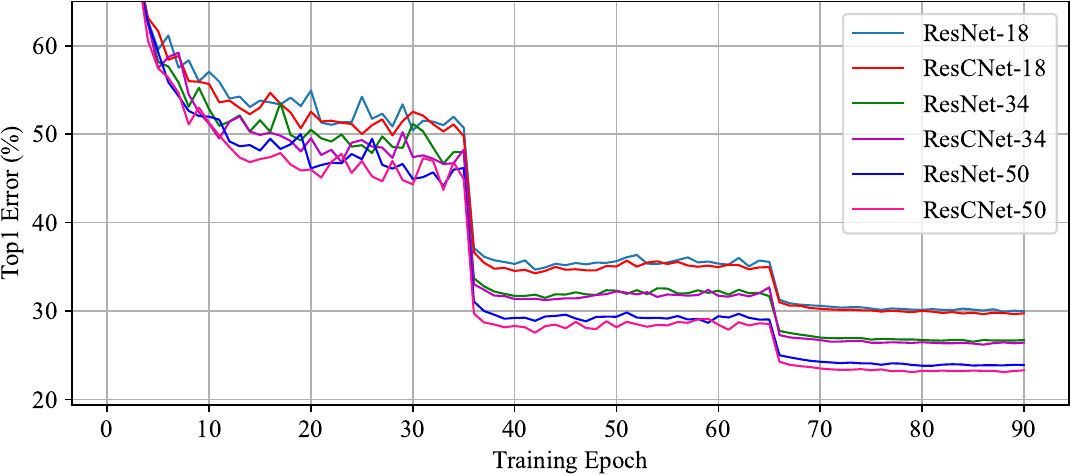}
    \caption{
        \textbf{Top-1 error of ImageNet using ResNet and ResCNet.}
        From the training curves, it can be observed that the error rate of the ResCNet model is lower than that of the ResNet model.
    }
    \label{fig:fig_err_imagenet}
\end{figure*}


We also validated the effectiveness of the ResCNet model on the ImageNet dataset, with the experimental results shown in Tab.~\ref{tab:imagenet_result} and Fig.~\ref{fig:fig_err_imagenet}. It can be observed that ResCNet effectively enhances the performance of the model.
Although SENet-50 was trained for 100 epochs, ResCNet-50 outperformed SENet-50 after training for just 90 epochs.
The training curves in Fig.~\ref{fig:fig_err_imagenet} demonstrate that the Top-1 error rate of the ResCNet model consistently remains below that of the ResNet model throughout the training process, indicating that the ResCNet model has better convergence capabilities.

\subsubsection{Visualization of Calibration Effect}

\begin{figure*}[!t]
    \centering
    \includegraphics[width=\textwidth]{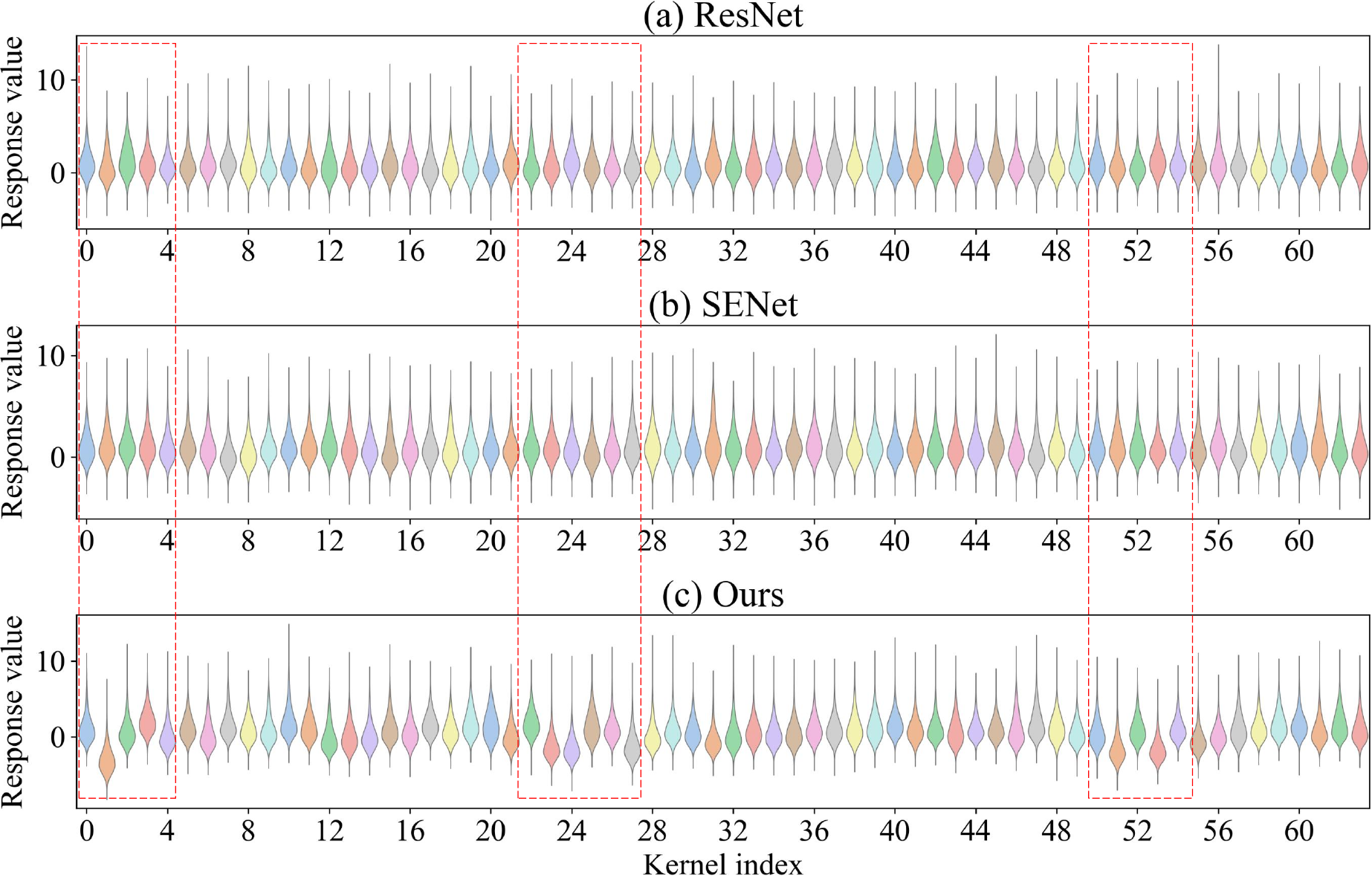}
    \caption{
        \textbf{Comparison of response value distributions.}
         We compared the response value distributions of the ResNet-32, SENet-32, and ResCNet-32 models on the CIFAR-100 dataset.
         The red dashed bounding boxes serve as visual indicators of the enhanced discriminability in neural feature responses.
    }
    \label{fig:ori_vs_ours_resnet32_violinplot}
\end{figure*}

We visualized the calibrated features and compared them with those of ResNet and SENet, as shown in Fig.~\ref{fig:ori_vs_ours_resnet32_violinplot}. 
It can be observed that the features generated by the method proposed in this paper exhibit stronger distinctiveness compared to other methods.
Although the SE layer can also adjust feature values, the extent of these adjustments is relatively small, and the distinction between different features is not pronounced.
Essentially, the SE layer primarily scales the features from the residual branch.
However, the approach proposed in this paper involves calculating additional calibration values for the features in the residual branch, which enhances the specificity among features, thereby providing stronger feature distinctiveness.

{\subsubsection{Exploration of ResCNet Performance Limits}

In this section, we explore the performance limits of the ResCNet model by employing several deep-learning tricks. 
We conducted experiments using the ResNet-50 and ResCNet-50 models on the CIFAR-100 dataset. The tricks used are as follows:
\begin{enumerate}
    \item Utilized the downsampling structure proposed by ResNet-D~\cite{he2019bag}.
    \item Increased training epochs to 300 and warm-up epochs to 20.
    \item AutoAugment (AA)~\cite{cubuk2019autoaugment}.
    \item Random Erasing ($p = 0.5$)~\cite{zhong2020random}.
    \item Mixup ($alpha = 1.0$)~\cite{zhang2017mixup}.
    \item Cutmix ($alpha = 1.0$)~\cite{yun2019cutmix}.
    \item Used a larger batch size of 256 without changing the learning rate.
    \item Distributed training (2 $\times$ RTX3090, the batch size of each GPU is 128).
\end{enumerate}

\begin{table*}[t]
    \centering
    {\begin{tabular}{@{}l|cc@{}}
\toprule
 & ResNet-50 & ResCNet-50 \\ \midrule
Baseline & 80.79 & 80.87 \\
+ResNet-D's Downsampling Block &80.91  &81.91  \\ 
+Long Training &82.19  &82.21  \\ 
+AutoAugment &83.14 &82.91 \\
+Random Erasing &83.58 &83.25 \\
+Mixup &84.60 &84.31 \\
+Cutmix &85.81 &85.73 \\
+Large BatchSize &85.84 &{86.17} \\
+Distributed training &85.92 &\textbf{86.31} \\
 
 \bottomrule
\end{tabular}}
    \caption{
        \textbf{CIFAR-100 result using ResNet-50 and ResCNet-50.}
        All scores are denoted in \%.
    }
    \label{tab:more_cifar100_result}
\end{table*}

We also used a technique similar to Random Erasing, known as Cutout~\cite{devries2017improved}, in our preliminary experiments. 
However, we found that Random Erasing yielded better results, approximately 0.4\% higher than Cutout.
Learning rate scaling did not yield good results. For instance, when the batch size was increased to 256 and the learning rate was adjusted to 0.2, the performance decreased by 0.3\%.
Overall, the hyperparameters we used were identified as optimal through extensive experimentation.
The experimental results are shown in Tab.~\ref{tab:more_cifar100_result}.

The experimental results show that the ResCNet-50 model performs excellently on the CIFAR-100 dataset by incorporating various deep-learning training tricks.
Compared to ResNet-50 ($85.92\%$), our proposed ResCNet-50 model achieves a superior performance of $86.31\%$.
To the best of our knowledge, we are the first to achieve over $86\%$ accuracy on the CIFAR-100 dataset with a 50-layer ResNet model variant without utilizing additional training data or pre-trained models\footnote{https://paperswithcode.com/sota/image-classification-on-cifar-100}, as of the time of submission. These results demonstrate the effectiveness of our method.}

\subsection{Ablation Study}\label{sec:ablation}
In this section, we primarily explore the performance differences resulting from the implementation of the RC Layer, 
as well as the impact of other calibration value integration methods on model performance.

\subsubsection{RC Layer Implementations}

In the specific implementation of the RC Layer, we employ three fully connected (FC) layers to fit the mean and standard deviation corresponding to each convolutional kernel.
In principle, using two FC layers is sufficient for learning the mean and standard deviation. However, employing two FC layers would significantly increase the model's parameter count. 
For instance, if the convolutional layer has $C$ channels, using two FC layers would increase the model's parameters by $2*C*C$. 
Conversely, with three FC layers, where the reduction factor in the middle layer is $r$, the total parameter count becomes $3*C*r$. 
Typically, $r$ is much smaller than $C$, thereby reducing the significant increase in model parameters.
Through this approach, we achieve feature calibration without a notable increase in the model's parameter count.

We compared the effects of different implementations of the RC Layer on the model. 
The model utilized was ResCNet-32, and the dataset employed was CIFAR-100. 
In the experiments, we observed that implementations with two FC layers and three FC layers produced similar results in terms of Top-1 accuracy. Consequently, we also present the Top-5 accuracy to offer a more precise comparison of model performance.
We also examined the impact of incorporating an activation function in the middle layer of the three FC layers implementation.
The experimental results are shown in Tab.\ref{tab:tab_rescnet32_fc}.

\begin{table*}[!t]
\centering
\resizebox{\textwidth}{!}{
\begin{tabular}{@{}ccccc@{}}
\toprule
\#FC &reduction & Activation layer & Top-1 Accuracy (\%) & Top-5 Accuracy (\%) \\ \midrule
2 & - & - &  \textbf{72.76}  & \textbf{92.67} \\
3 & 2 & - &  72.21 & 92.54 \\ 
3 & 4 & - &  \textbf{72.76}  & 92.54\\
3 & 4 & ReLU &  72.36 & 91.96 \\
3 & 4 & Sigmoid & 72.38 & 92.62 \\
\bottomrule
\end{tabular}
}
\caption{
    \textbf{The impact of different implementations of the RC Layer on model performance.}
    The dataset used is CIFAR-100, and the model is ResCNet-32.
    The first row of experimental results serves as the baseline.
    The \textbf{bold} indicates the best performance.
}
\label{tab:tab_rescnet32_fc}
\end{table*}

Experimental results show minimal differences in final model performance between implementations with two and three FC layers. 
Adjusting the reduction hyperparameter allows both implementations to achieve comparable performance levels. Given that the module proposed in this paper is an additional component designed not to increase the model's parameter count significantly, we consider these experimental results acceptable.
Experimental results also indicate that using an activation function does not enhance model performance.
We speculate that this may be because the activation function disrupts the module's learning process for the mean and variance of each neuron, despite potentially providing more nonlinear combinations.

\subsubsection{Other calibration value integration method}
\begin{figure*}[t]
    \centering
    \includegraphics[width=0.95\textwidth]{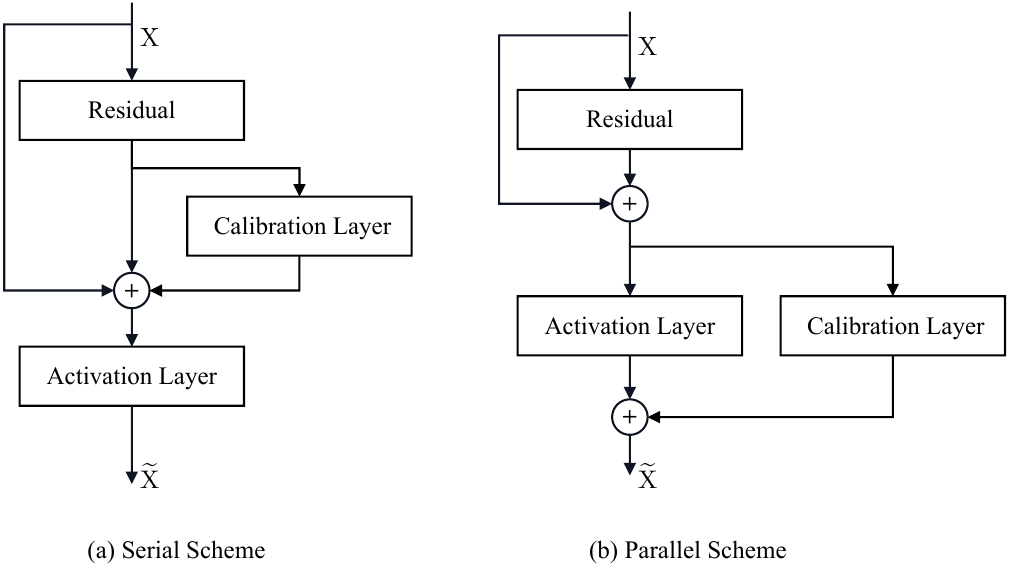}
    \caption{
        \textbf{Proposed integration schemes.}
        Sequential and parallel refer to the position of the calibration layer relative to the activation function layer.
        The original approach  (ResCNet) is termed ``serial scheme''.
    }
    \label{fig:integration_schemes}
\end{figure*}

In integrating the RC Layer, we position the calibration layer after the last BN in the ResNet block, just before the ReLU activation. 
We also experimented with placing the calibration layer alongside the ReLU and simplified the assumption of a Gaussian distribution to observe the resulting performance differences.
In other words, a duplicate of the input to the ReLU function is also passed to the adjustment calculation unit. Logically, both the adjustment calculation unit and the ReLU calculation unit receive the same input and compute in parallel.
We refer to this approach as parallel integration, while the original approach is termed serial integration.
``Serial" and ``Parallel" refer to their positioning relative to the original activation function.
The diagrams for the two integration methods are shown in Fig.~\ref{fig:integration_schemes}.

In the parallel scheme, we assume that the input features follow a standard Gaussian distribution with a mean of 0 and a standard deviation of 1. 
Additionally, we integrate this parallel computation into a design that resembles an activation function similar to ReLU. 
We name the activation function proposed in this paper as the Gaussian Calibration Linear Unit (GCLU).
In this case, the final activation value is the sum of the original response value and the calibration value.
Assuming the input is $a_{k}$, and $a_{k}\sim \mathcal{N}(0,1)$. The final response value is $f(a_{k})$, the formulaic expression is:

\begin{align}\label{eq:final_act_value}
f(a_{k}) &= \mathrm{ReLU}(a_{k}) + c_{k} \nonumber \\
         &= \mathrm{ReLU}(a_{k}) + a_{k} * w_{k} \nonumber \\
         &= \begin{cases}
                a_{k} * (2- \Phi(a_{k})), &a_{k} > 0 \\
                a_{k} * \Phi(a_{k}), &a_{k} \le 0
            \end{cases}.
\end{align}

\begin{table*}[!t]
\centering
\resizebox{\textwidth}{!}{
\begin{tabular}{@{}c|ccc|ccc@{}}
\toprule
Dataset & \multicolumn{3}{c|}{CIFAR-10} & \multicolumn{3}{c}{CIFAR-100} \\ \midrule
Activation layer & ReLU & GELU & GCLU (Ours) & ReLU & GELU & GCLU (Ours) \\ \midrule
AlexNet & 85.85 & 84.86 & \textbf{86.98} & 57.65 & 1.23 & \textbf{62.33}  \\
VGG-16 & 93.86 & 93.69 & \textbf{94.15} & 73.21 & 72.47 & \textbf{74.13}  \\
ResNet-8 & 88.14 & 88.37 & \textbf{88.38} & 60.73 & 60.61 & \textbf{60.82}  \\
ResNet-32 & 93.72 & \textbf{93.73} & 93.24 & \textbf{71.91} & 71.87 & 71.75  \\ \midrule
ViT-Tiny & 82.37 & \textbf{83.12} & {82.61} & 55.59 & \textbf{57.81} & {56.63}  \\
Swin-Tiny & \textbf{85.93} & 84.76 & {85.71} & \textbf{59.54} & 57.25 & {58.92}  \\
Cait-XXS & 84.28 & 84.14 & \textbf{84.46} & 59.16 & \textbf{59.77} & {59.67}  \\
\bottomrule
\end{tabular}}
\caption{
    \textbf{
    Accuracy performance of different activation functions.}
    The \textbf{bold} indicates the best performance.
    All scores are denoted in \%.
}
\label{tab:tab_act_result}
\end{table*}

Considering the generality of the activation function, we explored the performance of the parallel integration approach across more models utilizing the CIFAR-10 and CIFAR-100 datasets.
The evaluated model architectures include AlexNet, VGGNet, ResNet, Vision Transformer (ViT)~\cite{dosovitskiy2021vit}, Swin Transformer~\cite{liu2021swin}, and Cait~\cite{touvron2021going}.
The results are presented in Tab.~\ref{tab:tab_act_result}.

Experimental results indicate that the proposed activation function exhibits strong competitiveness relative to other mainstream activation functions. 
Additionally, issues with model non-convergence were observed when using AlexNet with GELU on the CIFAR-100 dataset. 
In contrast, the proposed method, similar to ReLU, did not encounter any issues with non-convergence. 
The results also demonstrate that the proposed activation function significantly enhances performance in models lacking Batch Normalization layers, such as AlexNet.
As network depth increases and Batch Normalization layers are introduced, the proposed activation function performs comparably to, or better than, other activation functions.
Overall, the activation function proposed in this paper demonstrates significant advantages in shallow convolutional neural network models and approaches to the performance of ReLU in deeper models, or GELU in visual transformers.

\begin{table*}[!t]
\centering
\resizebox{\textwidth}{!}{
\begin{tabular}{@{}c|cc|cc@{}}
\toprule
 \multirow{2}{*}{Approximation form} & \multicolumn{2}{c|}{CIFAR-10} & \multicolumn{2}{c}{CIFAR-100} \\ \cmidrule{2-5} 
 & \multicolumn{1}{l}{Accuracy (\%) } & Time (hour) & \multicolumn{1}{l}{Accuracy (\%)} & Time (hour) \\ \midrule
-       & 93.24 & 2.73 & 71.75 & 2.90 \\ 
Sigmoid & 93.35 & 2.43 & \textbf{72.32} & 2.40 \\ 
Tanh    & \textbf{93.57} & 2.67 &71.67 &2.73 \\ 
\bottomrule
\end{tabular}
}
\caption{
    \textbf{The influence of GCLU's approximation forms on ResNet-32's performance across CIFAR-10 and CIFAR-100 datasets.}
    "Sigmoid" represents the approximate form of Eq.~\ref{eq:sigmoid_approx}, and "Tanh" represents the approximate form of Eq.~\ref{eq:tanh_approx}. The first row of the experimental results represents the original results without using the approximate forms.
    The \textbf{bold} indicates the best performance.
}
\label{tab:tab_cdf_approx}
\end{table*}

Previous studies~\cite{hendrycks2016gaussian,page1977approximations,bowling2009logistic} have pointed out that there are two main approximation forms of the CDF of the standard normal distribution, namely:

\begin{equation}\label{eq:sigmoid_approx}
\Phi(x)\approx \sigma(1.702 x).
\end{equation}
and
\begin{equation}\label{eq:tanh_approx}
\Phi(x)\approx \frac{1}{2} \left[1 + \tanh\left(\sqrt{\frac{2}{\pi}}\left(x + 0.044715 x^3\right)\right)\right].
\end{equation}

We explored the impact of these two approximation forms on model performance using the ResNet-32 model on the CIFAR-10 and CIFAR-100 datasets.
The experimental results are presented in Tab.~\ref{tab:tab_cdf_approx}.
The experimental results show that the approximation forms of the function have a relatively minor impact on model performance, and the approximations can accelerate the training speed of the model in practice.
The approximate form of Eq.~\ref{eq:sigmoid_approx} has the shortest training time because it does not involve complex exponential, division, or square root operations.
Overall, the approximate forms of the GCLU activation function can reduce computational complexity, which is significant for promoting its application in mainstream neural networks.

\section{Conclusion}

In this paper, we propose a method to adjust neuron responses from a feature distribution perspective. This method involves adapting two trainable parameters, the mean and standard deviation for each neuron, which are used to calculate calibration values for their responses. These calibration values are subsequently added to the original responses. Based on these calibration values, we propose a plugin-based calibration module. This module is integrated into a modified ResNet architecture, termed Response Calibration Networks (ResCNet). Extensive experiments validate the effectiveness of the proposed response calibration technique. We hope this work will inspire future research on optimizing models from a distributional perspective.

\section*{Acknowledgment}
This work is supported by Ningbo Natural Science Foundation (2022J182) and Zhejiang Province ``Pioneering Soldier" and ``Leading Goose" R\&D Project under Grant 2023C01027. 

\bibliographystyle{elsarticle-num} 
\bibliography{ref}

\begin{thebibliography}{10}
\expandafter\ifx\csname url\endcsname\relax
  \def\url#1{\texttt{#1}}\fi
\expandafter\ifx\csname urlprefix\endcsname\relax\def\urlprefix{URL }\fi
\expandafter\ifx\csname href\endcsname\relax
  \def\href#1#2{#2} \def\path#1{#1}\fi

\bibitem{2012ImageNet}
A.~Krizhevsky, I.~Sutskever, G.~E. Hinton, Imagenet classification with deep
  convolutional neural networks, Advances in Neural Information Processing
  Systems 25 (2012).

\bibitem{2014Very}
K.~Simonyan, A.~Zisserman, Very deep convolutional networks for large-scale
  image recognition, arXiv preprint arXiv:1409.1556 (2014).

\bibitem{2014Going}
C.~Szegedy, W.~Liu, Y.~Jia, P.~Sermanet, S.~Reed, D.~Anguelov, D.~Erhan,
  V.~Vanhoucke, A.~Rabinovich, Going deeper with convolutions, in: Proceedings
  of the IEEE Conference on Computer Vision and Pattern Recognition, 2015, pp.
  1--9.

\bibitem{2016Deep}
K.~He, X.~Zhang, S.~Ren, J.~Sun, Deep residual learning for image recognition,
  in: Proceedings of the IEEE Conference on Computer Vision and Pattern
  Recognition, 2016, pp. 770--778.

\bibitem{2017Densely}
G.~Huang, Z.~Liu, L.~Van Der~Maaten, K.~Q. Weinberger, Densely connected
  convolutional networks, in: Proceedings of the IEEE Conference on Computer
  Vision and Pattern Recognition, 2017, pp. 4700--4708.

\bibitem{howard2017mobilenets}
A.~G. Howard, M.~Zhu, B.~Chen, D.~Kalenichenko, W.~Wang, T.~Weyand,
  M.~Andreetto, H.~Adam, Mobilenets: Efficient convolutional neural networks
  for mobile vision applications, arXiv preprint arXiv:1704.04861 (2017).

\bibitem{2016SqueezeNet}
F.~N. Iandola, S.~Han, M.~W. Moskewicz, K.~Ashraf, W.~J. Dally, K.~Keutzer,
  Squeezenet: Alexnet-level accuracy with 50x fewer parameters and <0.5 mb
  model size, arXiv preprint arXiv:1602.07360 (2016).

\bibitem{guariglia2018harmonic}
E.~Guariglia, Harmonic sierpinski gasket and applications, Entropy 20~(9)
  (2018) 714.

\bibitem{guariglia2019primality}
E.~Guariglia, Primality, fractality, and image analysis, Entropy 21~(3) (2019)
  304.

\bibitem{guariglia2021fractional}
E.~Guariglia, Fractional calculus, zeta functions and shannon entropy, Open
  Mathematics 19~(1) (2021) 87--100.

\bibitem{yang2019hyperspectral}
L.~Yang, H.~Su, C.~Zhong, Z.~Meng, H.~Luo, X.~Li, Y.~Y. Tang, Y.~Lu,
  Hyperspectral image classification using wavelet transform-based smooth
  ordering, International Journal of Wavelets, Multiresolution and Information
  Processing 17~(06) (2019) 1950050.

\bibitem{basha2020impact}
S.~S. Basha, S.~R. Dubey, V.~Pulabaigari, S.~Mukherjee, Impact of fully
  connected layers on performance of convolutional neural networks for image
  classification, Neurocomputing 378 (2020) 112--119.

\bibitem{qureshi2022neurocomputing}
K.~N. Qureshi, O.~Kaiwartya, G.~Jeon, F.~Piccialli, Neurocomputing for internet
  of things: object recognition and detection strategy, Neurocomputing 485
  (2022) 263--273.

\bibitem{zhou2024patchdetector}
L.~Zhou, S.~Zhang, T.~Qiu, W.~Xu, Z.~Feng, M.~Song, Patchdetector: Pluggable
  and non-intrusive patch for small object detection, Neurocomputing 589 (2024)
  127715.

\bibitem{mallat1989theory}
S.~G. Mallat, A theory for multiresolution signal decomposition: the wavelet
  representation, IEEE Transactions on Pattern Analysis and Machine
  Intelligence 11~(7) (1989) 674--693.

\bibitem{zheng2019framework}
X.~Zheng, Y.~Y. Tang, J.~Zhou, A framework of adaptive multiscale wavelet
  decomposition for signals on undirected graphs, IEEE Transactions on Signal
  Processing 67~(7) (2019) 1696--1711.

\bibitem{guido2017effectively}
R.~C. Guido, Effectively interpreting discrete wavelet transformed signals,
  IEEE Signal Processing Magazine 34~(3) (2017) 89--100.

\bibitem{guariglia2016fractional}
E.~Guariglia, S.~Silvestrov, Fractional-wavelet analysis of positive definite
  distributions and wavelets on d'(c), in: Engineering mathematics II,
  Springer, 2016, pp. 337--353.

\bibitem{MENG2023PINN}
Z.~Meng, Q.~Qian, M.~Xu, B.~Yu, A.~R. Yıldız, S.~Mirjalili, Pinn-form: A new
  physics-informed neural network for reliability analysis with partial
  differential equation, Computer Methods in Applied Mechanics and Engineering
  414 (2023) 116172.

\bibitem{raissi2024forward}
M.~Raissi, Forward--backward stochastic neural networks: deep learning of
  high-dimensional partial differential equations, in: Peter Carr
  Gedenkschrift: Research Advances in Mathematical Finance, World Scientific,
  2024, pp. 637--655.

\bibitem{muhammad2018early}
K.~Muhammad, J.~Ahmad, S.~W. Baik, Early fire detection using convolutional
  neural networks during surveillance for effective disaster management,
  Neurocomputing 288 (2018) 30--42.

\bibitem{gupta2021deep}
A.~Gupta, S.~Watson, H.~Yin, Deep learning-based aerial image segmentation with
  open data for disaster impact assessment, Neurocomputing 439 (2021) 22--33.

\bibitem{yu2021convolutional}
H.~Yu, L.~T. Yang, Q.~Zhang, D.~Armstrong, M.~J. Deen, Convolutional neural
  networks for medical image analysis: state-of-the-art, comparisons,
  improvement and perspectives, Neurocomputing 444 (2021) 92--110.

\bibitem{niyas2022medical}
S.~Niyas, S.~Pawan, M.~A. Kumar, J.~Rajan, Medical image segmentation with 3d
  convolutional neural networks: A survey, Neurocomputing 493 (2022) 397--413.

\bibitem{ioffe2015bn}
S.~Ioffe, C.~Szegedy, Batch normalization: Accelerating deep network training
  by reducing internal covariate shift, in: International Conference on Machine
  Learning, pmlr, 2015, pp. 448--456.

\bibitem{hu2018senet}
J.~Hu, L.~Shen, G.~Sun, Squeeze-and-excitation networks, in: Proceedings of the
  IEEE Conference on Computer Vision and Pattern Recognition, 2018, pp.
  7132--7141.

\bibitem{chen2024symbolic}
X.~Chen, C.~Liang, D.~Huang, E.~Real, K.~Wang, H.~Pham, X.~Dong, T.~Luong,
  C.-J. Hsieh, Y.~Lu, et~al., Symbolic discovery of optimization algorithms,
  Advances in neural information processing systems 36 (2024).

\bibitem{dou2024gbrun}
Z.-C. Dou, S.-C. Chu, Z.~Zhuang, A.~R. Yildiz, J.-S. Pan, Gbrun: A gradient
  search-based binary runge kutta optimizer for feature selection, Journal of
  Internet Technology 25~(3) (2024) 341--353.

\bibitem{ba2016ln}
J.~L. Ba, J.~R. Kiros, G.~E. Hinton, Layer normalization, arXiv preprint
  arXiv:1607.06450 (2016).

\bibitem{hendrycks2016gaussian}
D.~Hendrycks, K.~Gimpel, Gaussian error linear units (gelus), arXiv preprint
  arXiv:1606.08415 (2016).

\bibitem{li2019sknet}
X.~Li, W.~Wang, X.~Hu, J.~Yang, Selective kernel networks, in: Proceedings of
  the IEEE Conference on Computer Vision and Pattern Recognition, 2019, pp.
  510--519.

\bibitem{erdacs2023optimum}
M.~U. Erda{\c{s}}, M.~Kopar, B.~S. Yildiz, A.~R. Yildiz, Optimum design of a
  seat bracket using artificial neural networks and dandelion optimization
  algorithm, Materials Testing 65~(12) (2023) 1767--1775.

\bibitem{sait2024optimal}
S.~M. Sait, P.~Mehta, A.~R. Y{\i}ld{\i}z, B.~S. Y{\i}ld{\i}z, Optimal design of
  structural engineering components using artificial neural network-assisted
  crayfish algorithm, Materials Testing (2024).

\bibitem{wu2018gn}
Y.~Wu, K.~He, Group normalization, in: Proceedings of the European Conference
  on Computer Vision (ECCV), 2018, pp. 3--19.

\bibitem{ulyanov2016instance}
D.~Ulyanov, A.~Vedaldi, V.~Lempitsky, Instance normalization: The missing
  ingredient for fast stylization, arXiv preprint arXiv:1607.08022 (2016).

\bibitem{woo2018cbam}
S.~Woo, J.~Park, J.-Y. Lee, I.~S. Kweon, Cbam: Convolutional block attention
  module, in: Proceedings of the European Conference on Computer Vision (ECCV),
  2018, pp. 3--19.

\bibitem{li2021involution}
D.~Li, J.~Hu, C.~Wang, X.~Li, Q.~She, L.~Zhu, T.~Zhang, Q.~Chen, Involution:
  Inverting the inherence of convolution for visual recognition, in:
  Proceedings of the IEEE Conference on Computer Vision and Pattern
  Recognition, 2021, pp. 12321--12330.

\bibitem{vaswani2021scaling}
A.~Vaswani, P.~Ramachandran, A.~Srinivas, N.~Parmar, B.~Hechtman, J.~Shlens,
  Scaling local self-attention for parameter efficient visual backbones, in:
  Proceedings of the IEEE Conference on Computer Vision and Pattern
  Recognition, 2021, pp. 12894--12904.

\bibitem{krizhevsky2009learning}
A.~Krizhevsky, G.~Hinton, Learning multiple layers of features from tiny
  images, Master's thesis, Department of Computer Science, University of
  Toronto (2009).

\bibitem{kingma2013auto}
D.~Kingma, Auto-encoding variational bayes, arXiv preprint arXiv:1312.6114
  (2013).

\bibitem{netzer2011reading}
Y.~Netzer, T.~Wang, A.~Coates, A.~Bissacco, B.~Wu, A.~Y. Ng, et~al., Reading
  digits in natural images with unsupervised feature learning, in: NIPS
  workshop on deep learning and unsupervised feature learning, Granada, Spain,
  2011, p.~7.

\bibitem{deng2009imagenet}
J.~Deng, W.~Dong, R.~Socher, L.-J. Li, K.~Li, L.~Fei-Fei, Imagenet: A
  large-scale hierarchical image database, in: 2009 IEEE Conference on Computer
  Vision and Pattern Recognition, Ieee, 2009, pp. 248--255.

\bibitem{loshchilov2016sgdr}
I.~Loshchilov, F.~Hutter, Sgdr: Stochastic gradient descent with warm restarts,
  arXiv preprint arXiv:1608.03983 (2016).

\bibitem{he2019bag}
T.~He, Z.~Zhang, H.~Zhang, Z.~Zhang, J.~Xie, M.~Li, Bag of tricks for image
  classification with convolutional neural networks, in: Proceedings of the
  IEEE Conference on Computer Vision and Pattern Recognition, 2019, pp.
  558--567.

\bibitem{cubuk2019autoaugment}
E.~D. Cubuk, B.~Zoph, D.~Mane, V.~Vasudevan, Q.~V. Le, Autoaugment: Learning
  augmentation strategies from data, in: Proceedings of the IEEE Conference on
  Computer Vision and Pattern Recognition, 2019, pp. 113--123.

\bibitem{zhong2020random}
Z.~Zhong, L.~Zheng, G.~Kang, S.~Li, Y.~Yang, Random erasing data augmentation,
  in: Proceedings of the AAAI conference on Artificial Intelligence, 2020, pp.
  13001--13008.

\bibitem{zhang2017mixup}
H.~Zhang, M.~Cisse, Y.~N. Dauphin, D.~Lopez-Paz, mixup: Beyond empirical risk
  minimization, arXiv preprint arXiv:1710.09412 (2017).

\bibitem{yun2019cutmix}
S.~Yun, D.~Han, S.~J. Oh, S.~Chun, J.~Choe, Y.~Yoo, Cutmix: Regularization
  strategy to train strong classifiers with localizable features, in:
  Proceedings of the IEEE International Conference on Computer Vision, 2019,
  pp. 6023--6032.

\bibitem{devries2017improved}
T.~DeVries, G.~W. Taylor, Improved regularization of convolutional neural
  networks with cutout, arXiv preprint arXiv:1708.04552 (2017).

\bibitem{dosovitskiy2021vit}
A.~Dosovitskiy, L.~Beyer, A.~Kolesnikov, D.~Weissenborn, X.~Zhai,
  T.~Unterthiner, M.~Dehghani, M.~Minderer, G.~Heigold, S.~Gelly, et~al., An
  image is worth 16x16 words: Transformers for image recognition at scale,
  arXiv preprint arXiv:2010.11929 (2020).

\bibitem{liu2021swin}
Z.~Liu, Y.~Lin, Y.~Cao, H.~Hu, Y.~Wei, Z.~Zhang, S.~Lin, B.~Guo, Swin
  transformer: Hierarchical vision transformer using shifted windows, in:
  Proceedings of the IEEE international conference on computer vision, 2021,
  pp. 10012--10022.

\bibitem{touvron2021going}
H.~Touvron, M.~Cord, A.~Sablayrolles, G.~Synnaeve, H.~J{\'e}gou, Going deeper
  with image transformers, in: Proceedings of the IEEE international conference
  on computer vision, 2021, pp. 32--42.

\bibitem{page1977approximations}
E.~Page, Approximations to the cumulative normal function and its inverse for
  use on a pocket calculator, Journal of the Royal Statistical Society Series
  C: Applied Statistics 26~(1) (1977) 75--76.

\bibitem{bowling2009logistic}
S.~R. Bowling, M.~T. Khasawneh, S.~Kaewkuekool, B.~R. Cho, A logistic
  approximation to the cumulative normal distribution, Journal of Industrial
  Engineering and Management 2~(1) (2009) 114--127.

\end{thebibliography}

\end{document}